\documentclass{article}

\usepackage{microtype}
\usepackage{graphicx}
\usepackage{subcaption}
\usepackage{booktabs} 

\usepackage[accepted]{icml2019}

\newcommand{\papertitle}{Robust conditional GANs under missing or uncertain labels}
\newcommand{\runningtitle}{\papertitle}

\icmltitlerunning{\runningtitle}

\usepackage{amssymb,amsfonts,amsmath,amscd,dsfont,mathrsfs}
\usepackage{pgf,pgfarrows,pgfnodes}
\usepackage{multicol, algorithmic,multirow}

\usepackage{caption}
\usepackage{tikz}
\usepackage{lipsum}
\usepackage{enumitem}

\usepackage{color}

\usepackage{amsmath,amsfonts, amssymb}
\usepackage{amsthm}
\usepackage{mathrsfs}

\usepackage{dsfont} 
\usepackage{xspace}

\newtheorem{theorem}{Theorem}
\newtheorem{corollary}{Corollary}

\newtheorem{assume}{Assumption}

\newcommand{\prob}[1]{ \mathbb{P}(#1) }

\newcommand{\reals}{\mathbb{R}}

\newcommand{\ones}{\mathds{1}}
\newcommand{\identy}{\mathbf{I}}
\newcommand{\diag}{\mathbf{diag}}

\newcommand{\abs}[1]{\left| #1 \right|}

\newcommand{\maxnorm}[1]{\vertiii{#1}_{\infty}}

\newcommand{\vopsymb}{\circ}
\newcommand{\vop}[2]{#1 \vopsymb #2}

\newcommand{\cF}{\mathcal{F}}

\newcommand{\cG}{\mathcal{G}}

\newcommand{\cX}{\mathcal{X}}

\newcommand{\ty}{\widetilde{y}}
\newcommand{\tY}{\widetilde{Y}}
\newcommand{\D}{D}

\newcommand{\ny}{m}

\newcommand{\nndff}[3]{d_{#1} (#2, #3 )}
\newcommand{\expectD}[2]{\underset{#1}{\mathbb{E}}\left[ #2 \right]}

\newcommand{\xy}[1]{#1_{X, Y}}
\newcommand{\xty}[1]{#1_{X, \tY}}
\newcommand{\tm}{\widetilde{m}}

\newcommand{\tP}{\widetilde{P}}
\newcommand{\tQ}{\widetilde{Q}}
\newcommand{\htP}{\widetilde{P}_n}
\newcommand{\htQ}{\widetilde{Q}_n}

\newcommand{\vertiii}[1]{{\vert\kern-0.25ex\vert\kern-0.25ex\vert #1 
		\vert\kern-0.25ex\vert\kern-0.25ex\vert}}

\newcommand{\tv}[2]{d_{\rm TV}\left(#1, #2\right)}
\usepackage{mathtools}
\DeclarePairedDelimiterX{\infdivx}[2]{(}{)}{%
	#1\;\delimsize\|\;#2%
}

\newcommand{\js}{d_{\rm JS}\infdivx*}

\newcommand{\confus}{C}

\renewcommand\footnotemark{}

\begin{document}

\twocolumn[
\icmltitle{\papertitle}



\icmlsetsymbol{equal}{*}

\begin{icmlauthorlist}
\icmlauthor{Kiran Koshy Thekumparampil}{uiuc}
\icmlauthor{Sewoong Oh}{uw}
\icmlauthor{Ashish Khetan}{aws}
\end{icmlauthorlist}

\icmlaffiliation{uiuc}{
	University of Illinois at Urbana-Champaign}
\icmlaffiliation{uw}{
	University of Washington, Seattle}
\icmlaffiliation{aws}{
	Amazon, New York
}

\icmlcorrespondingauthor{Kiran K. Thekumparampil}{thekump2@illinois.edu}

\icmlkeywords{conditional GAN,semi-supervised,few labels,complementary labels, robust}

\vskip 0.3in
]



\printAffiliationsAndNotice{}  


\begin{abstract}
Matching the performance of conditional Generative Adversarial Networks 
with little supervision is an important task, especially in venturing into new domains. 
We design a new training algorithm,  
which is robust to missing or ambiguous labels. 
The main idea is to intentionally corrupt the 
labels of generated examples to match the statistics of the real data, 
and have a discriminator process the real and generated examples with corrupted labels.
We showcase the robustness of this proposed approach both theoretically and empirically. 
We show that minimizing the proposed loss is equivalent to minimizing 
true divergence between real and generated data up to a multiplicative factor, 
and characterize this multiplicative factor as a function of the statistics of the uncertain labels.
Experiments on MNIST dataset demonstrates that proposed architecture is 
able to achieve high accuracy in generating examples faithful to the class even with only a few examples per class. 
\end{abstract}

\section{Introduction} \label{sec:intro}
Conditional GAN (cGAN) has been applied to several domains for various tasks, 
such as improving image quality, reinforcement learning, and category transformation \cite{mirza2014conditional,LTH16,ZPI17,OOS16}.
As opposed to a standard GAN, a conditional GAN is trained using labeled samples which provide additional useful information, which could be utilized to generate better quality samples \cite{brock2018large}. 
However, it is costly to obtain accurate class labels for all the samples. 
Instead, we might choose to collect  accurate labels for a few examples, 
and either leave most examples without labels 
or find cheaper ways to collect less accurate labels. 
In this paper, 
we consider a class of such economically collected labels, which we call {\em uncertain} labels. 
We provide a robust cGAN architecture with finite sample performance guarantees and empirically verify the its performance for the case of missing labels.

\noindent
{\bf Notation.} $[m] = \{1, 2, \cdots, m\}$, $\ones_k \in \reals^k$ is the all ones vector, $e_k$ is the $k$-th standard basis vector (with appropriate dimensions), $\identy_{k} \in \reals^{k \times k}$ is the identity matrix, ${\diag}(v)$ denotes a diagonal matrix with $v$ as the diagonal, and for $A \in \reals^{k \times k}$ we define $\| A\|_\infty = \max_{ \in [k]} \sum_{j \in [k]} |A_{ij}|$.


\noindent{\bf Uncertainty model.}
Let $x \in \cX$ be a data point having a {\em true label} $y \in [m]$ drawn from a joint distribution $P_{X, Y}$. 
We consider a semi-supervised setting, where 
we observe only a few examples with correct labels. 
The remaining examples have 
labels that are corrupted by uncertainty. 
Concretely, 
there is an additional set of $\tm$ labels $\{m+1,m+2,\ldots,m+\tm\}$. 
Having an example $x_i$ with {\em observed} label $\ty_i=m+1$, for example, means we are uncertain about the 
true label $y_i$, but we have some information about it according to the observed label $m+1$. 
A common example is the standard semi-supervised setting where 
$\tm=1$, and the class $m+1$ indicates that the label is missing. 
Another example is when the crowd is asked to give a membership, instead of a definite class, 
where a label $\ty_i=m+1$ might mean that 
the example $x_i$ has one of three labels $\{1,5,8\}$ but we are uncertain about which one.
We refer to the set of true labels $\{1,\ldots,m\}$ as {\em class labels} and 
the set of corrupted labels $\{m+1,\ldots,m+\tm\}$ as {\em uncertain labels}. 

We assume that each data point is corrupted independently and with a certain probability conditioned on the true label by an erasure channel. 
Formally, each $\ty_i$ is drawn according to a confusion matrix $C\in\reals^{(m+\tm)\times(m+\tm)}$ where 
$C_{ju} = \prob{\tY=u|Y=j}$.
Unlike the standard noisy label setting, we only consider uncertain labels; 
if you observe one of the class labels, then you are certain that it is the correct label. 
Otherwise, each uncertain label has  an uncertainty set that the label could have been generated from. 
Formally,
an uncertain label $u$ is parameterized by a vector $\alpha_u \in [0,1]^{m+\tm} $, where 
$\alpha_{ui}  = \prob{\tY=u\,|\,Y=i}$ if $i\in[m]$ and $\alpha_{ui} = 0$ if $i\in \{m+1,\ldots,m+\tm\}$.
The zeros follow from the fact that the true label cannot be an uncertain label. 
It immediately follows that  $\prob{\tY=i\,|\,Y=i} = 1 - \sum_{u = m}^{m+\tm}  \alpha_{ui}$.  
Under such an {\em uncertainty} model, the confusion matrix can be written as 
\begin{align}\label{eq:confusion_matrix}
C = \diag\Big(\,\ones_{m+n} - \sum_{u=m+1}^{m+\tm} \alpha_u\,\Big)+ \sum_{u =  m+1}^{m+\tm} {\alpha}_u e_{u}^T\;.
\end{align}

This captures a variety of label corruption models:
\begin{enumerate}[label=(\alph*)]
	\item {\bf Missing labels}: If $\alpha$ portion of the samples have their labels missing, then we can can incorporate the missing labels into our model as the uncertain class $u$, with $\alpha_{u} = [\alpha \ones^T_{m} \;\; 0]^T$.
	\item {\bf Complementary labels} \cite{ishida2017learning}: A complementary label specifies that a sample does not belong to a particular class. Let all samples from each class $y$ are assigned a complimentary label uniformly at random from $[m] \setminus \{y\}$. Then the complimentary label which specifies the exclusion from class $y$ could be denoted by the uncertain label $u_y$ with $\alpha_{u_y} = [(\ones_{m} - e_y)^T \;\; 0]^T/(m-1)$.
	\item {\bf Group (membership) labels}: Group label specify if a sample belongs to a subset of classes or not. For example, if the original classes are: {\em car, bus, horse, cat}, then 
	we could divide them into two super group labels: {\em automobile, animal}.
	It can easily be shown that this is a special case of our uncertainty model.
\end{enumerate}

\noindent
{\bf Contribution.}
In this paper, we design a new adversarial training of deep generative models,  
which is robust against uncertainty models discussed above. 
The main idea is to intentionally corrupt the 
label of generated examples, and have a discriminator distinguish the 
real and generated $(x,\ty)$: 
data example $x$ and corrupted label $\ty$, jointly. 
We showcase the robustness of this proposed approach both theoretically and empirically. 
First, we show that minimizing the proposed loss is equivalent to minimizing 
true divergence between real and generated $(x,y)$ up to a multiplicative factor 
(Theorems~\ref{thm:prob_dist_ub_lb} and \ref{thm:gen_ub_lb}). This multiplicative factor characterizes how  
the performance depends on the uncertainty parameters $\alpha_u$'s. 
We further provide sample complexity of achieving the same guarantee in Theorem~\ref{thm:nn_sample}. 
Experiments on MNIST dataset demonstrates that proposed architecture is 
able to achieve 97\% accuracy in generating examples faithful to the class even with only a few labeled examples per digit. 

\noindent
{\bf Related work.}
As semi-supervised learning was one of the initial motivations of training deep generative models, 
training a GAN with a few labeled examples has been an important topic of interest. 
\citet{SGZ16} used (unconditional) GAN as a proxy for training a semi-supervised classifier. 
\citet{sricharan2017semi} proposed training conditional GANs, but using two discriminators: 
one for distinguishing real and generated $x$ and another for distinguishing real and generated $(x,y)$.  
\citet{lucic2019high} proposed training a conditional GAN 
by first training a classifier using off-the-shelf semi-supervised techniques, 
and then using this to complete the missing labels with the help of an additional self-supervised discriminator.
 They get high-fidelity images, trained on ImageNet data. 
\citet{xu2019generative} studied training classifiers under complementary labels. 

For the rest of the manuscript, if $P_{X, Y}$ is the distribution of the true labeled data, then $\tP_{X, \tY}$ denotes the distribution of the corrupt labeled data corrupted by the the uncertainty model represented by
$C$ in eq.~\eqref{eq:confusion_matrix}.

\section{Robust cGAN (RCGAN) architecture}

We suppose that we know the confusion matrix $C$. 
It is easy to estimate, for example, 
when the only uncertain label is the missing label (assuming known marginal $P_Y$ as usual for cGANs). 
We propose the robust conditional GAN (RCGAN) architecture, inspired from the RCGAN for noisy labeled data \cite{thekumparampil2018robustness}. 
RCGAN uses the following adversarial loss $L_{\rm }(D, G)$:
\begingroup
\allowdisplaybreaks
\begin{align} \label{eq:loss_adv}
L_{\rm }(D, G) = &\expectD{(x,\ty) \sim \tP_{X,\tY}}{\phi\left(\D(x,\ty)\right)} + \nonumber \\ &\expectD{\substack{z \sim N, \, y \sim P_{Y} \\ \ty|y \sim \confus_y}}{\phi\left(1- \D(G(z;y), \ty)\right)} \,, 
\end{align}
\endgroup
where $D: \cX \times \reals^{m+\tm} \to \reals\,$ is the conditional discriminator, $G : \mathcal{Z} \times \reals^{m+\tm} \to \cX $ is the conditional generator, $\mathcal{Z}$ is the domain of input latent $z$,
and $\phi$ and $\ell$ are some loss functions. The discriminator and generator update steps (in  order) are given by:
$
\max_{D \in \cF} \; L_{\rm }(D, G) 
\, \text{ and,   } 
\min_{G \in \cG}  \; L_{\rm }(D, G) 
\,,
$
where $\cF$ is the family of conditional discriminators, and $\cG$ is the family of conditional generators. Note, that the generated sample $G(z; y)$ is a function of latent vector $z$ with distribution $N$ and is conditioned on the true label $y$ generated according true marginal $P_Y$.

\noindent
The first expectation is estimated with the corrupted real labeled samples, whose distribution is $\tP_{X, \tY}$. 
The second expectation is taken over the generator input latent ($z$) distribution $N$, the true class marginal $P_Y$, and the distribution, $C_y$ ($y$-th row of the confusion matrix), of the corrupted label $\ty$ given the true label $y$. That is, the true label $y$, of the generator samples are artificially corrupted to $\ty$, by the same uncertainty model which corrupted the real data. Thus the discriminator $D$ computes a distance between  the corrupted real labeled distribution $\tP_{X,\tY}$ and the corrupted generated labeled distribution, denoted by $\tQ_{X, \tY}$ and in Section \ref{sec:theory} we reason why minimizing this distance would minimize the distance between the true real and generated distributions $P_{X, Y}, Q_{X, Y}$. For this loss we use the projection discriminator \cite{miyato2018cgans} of the form discribed in Section \ref{sec:theory}.

\begin{figure}[h]
	\centering
	\includegraphics[width=.45\textwidth]{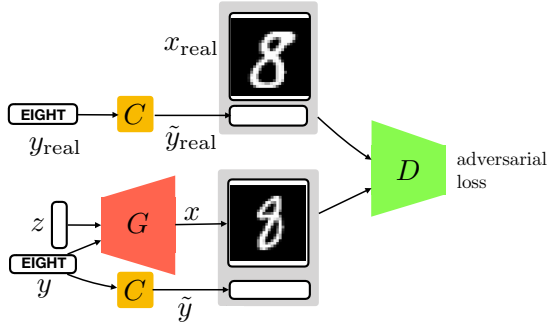}
	\vspace{-0.5cm}
	\caption{
		RCGAN: The output $x$ of the generator $G$ is paired with an uncertain label $\ty$, which is corrupted by the same {\em uncertainty} model, $C$, which corrupted the uncertain real label $\ty_{\rm real}$. The discriminator $D$ estimates whether a given labeled sample is coming from the real data $(x_{\rm real},\tilde{ y}_{\rm real})$ or the generated data $(x,\tilde{y})$.}
	\label{fig:RCGAN}
\end{figure}

\subsection{Theoretical Analysis of RCGAN}\label{sec:theory}
We see that our proposed RCGAN loss $L_{\rm }(D, G)$ \eqref{eq:loss_adv} minimizes a divergence, $d_{\cF}(\tP, \tQ)$ between the distribution, $\tP$, of the given corrupt real samples and distribution, $\tQ$, of the generated samples whose labels are artificially corrupted by the same uncertainty model, $C$,  which corrupted the real data, where,
\begin{align} \label{eq:nnd}
d_{\cF}(\tP, \tQ) = &\max_{D \in \cF} \expectD{(x,\ty) \sim \tP_{X,\tY}}{\phi\left(\D(x,\ty)\right)} + \nonumber \\ &\expectD{(x,\ty) \sim \tQ_{X,\tY}}{\phi\left(1 - \D(x,\ty)\right)} \,.
\end{align}
When $\cF$ is the set of all functions with range $[0, 1]$, this divergence reduces to the standard GAN losses: (a) the total variation distance $d_{\rm TV}(\tP, \tQ) \triangleq \max_{S \in \cX \times [m]} \{\tP(S) - \tQ(S)\}$ when $\phi(x)=x$ (up to some scaling and shifting) and (b) the Jensen-Shannon divergence $d_{\rm JS}(\tP, \tQ) \triangleq (1/2)(D_{\rm KL}(\tP || (\tP + \tQ)/2) + D_{\rm KL}(\tQ || (\tP + \tQ)/2))$ when $\phi(x)=\log x$ ($D_{\rm KL}$ is the Kullback-Leibler divergence). Next, we provide some approximation guarantees on these divergences to motivate our proposed architecture which corrupts the generated samples.
\begin{theorem} \label{thm:prob_dist_ub_lb}
	Let $\xy{P}$ and $\xy{Q}$ be two distributions on $\cX \times [\ny]$. 
	Let $\xty{\tP}$ and $\xty{\tQ}$ 
	be the corresponding  distributions when samples from $P, Q$ 
	are passed through the erasure channel given by the confusion matrix $\confus \in \reals^{(\ny+\tm) \times (\ny+\tm)}$ (eq. \eqref{eq:confusion_matrix}).
	If $\confus$ is full-rank ($\sum_u \alpha_u \prec \ones$), and $\kappa_\alpha =  \frac{1}{1-\|\sum_u \alpha_{u}\|_\infty} = \max_{i \in [m]} \frac{1}{1-\sum_u \alpha_{ui}}$, we get,
	\begin{align}
	\tv{\tP}{\tQ} \;\;\leq  &\;\;\tv{P}{Q}\;\;  \leq \;\; \kappa_\alpha \, \tv{\tP}{\tQ} \;,\text{ and } \label{eq:tv_ub_lb} \\
	\js{\tP}{\tQ}  \;\; \leq  &\;\;\js{P}{Q}\;\; \leq\;\;  \kappa_\alpha  \sqrt{8\, \js{\tP}{\tQ}}  \;.
	\label{eq:js_ub_lb}
	\end{align}
\end{theorem}
A proof is provided in Appendix \ref{app:proof_prob_dist_ub_lb}. These bounds imply that minimizing the divergences between the corrupt distributions $\tP, \tQ$ will minimize the divergence between the true distributions $P, Q$. However, these divergences do not generalize under finite sample assumptions, therefore we study a more practical GAN loss, called the {\em neural network distance} which could generalize \cite{arora2017generalization}. We say that the divergence $d_{\cF}(\tP, \tQ)$ is a neural network distance when the class of discriminators $\cF$ is parameterized by a finite set of variables (like in a neural network). For simplicity, we assume that $\phi(x)=x$. 

For deriving similar approximation bounds as in Theorem \ref{thm:prob_dist_ub_lb}, we make the simple Assumption \ref{assume:invariance} (Appendix \ref{app:invariance}) on the discriminator function class $\cF$ \cite{thekumparampil2018robustness}.  It is easy to show that the state-of-the-art {\em projection discriminator}  \cite{miyato2018cgans}, $\D_{V,v,\theta}(x,y)$ will satisfy the assumption, when it has the following form:
\begin{align*}
\D_{V,v,\theta}(x,y) \;\; = \;\;  {\rm vec}(y)^T \, V\,  \psi(x;\theta)  \,+\,  v^T\,\psi'(x;\theta)\;, 
\end{align*}
where ${\rm vec}(y)^T = [{\mathbb I}_{y=1} , \ldots, {\mathbb I}_{y=m+\tm} ]$, $\psi, \psi' \in \reals^{d}$ are any neural networks parameterized by $\theta$, $v \in \reals^{d}$, and $V \in \reals^{(m+\tm) \times d}$ such that $V \in \{ V \,|\, \max_{ij} |V| \leq 1 \}$  \cite{thekumparampil2018robustness}. This constraint on $V$ can be easily implemented through weight clipping. Next we show that, the neural network distance satisfies similar guarantees as the total variation distance.

\begin{theorem} \label{thm:gen_ub_lb}
	Under the same assumptions as in Theorem \ref{thm:prob_dist_ub_lb}, if a class of functions $\cF$ satisfies Assumption \ref{assume:invariance}, then 
	\begin{align}
	\nndff{\cF}{\tP}{\tQ} \;\; \leq \;\; \nndff{\cF}{P}{Q} \;\; \leq \;\;  \kappa_\alpha'  \nndff{\cF}{\tP}{\tQ} \;,
	\label{eq:main1}
	\end{align}
	where $\kappa_\alpha' =  \frac{1+\|\sum_u \alpha_{u}\|_\infty}{1-\|\sum_u \alpha_{u}\|_\infty} = \max_{i \in [m]} \frac{1+\sum_u \alpha_{ui}}{1-\sum_u \alpha_{ui}}$.
\end{theorem}
Similar to that of Theorem \ref{thm:prob_dist_ub_lb}, a proof of the above theorem follows from \citet[Theorem 2 ]{thekumparampil2018robustness}. This justifies the proposed RCGAN architecture to learn the true conditional distribution from corrupted labels. However, in practice, we observe only $n$ samples from each of the distributions $\tP$, $\tQ$, and we minimize the empirical divergence $d_{\cF}(\tP_n, \tQ_n)$ between the empirical distributions, $\tP_n$, $\tQ_n$ of these samples \cite{thekumparampil2018robustness}. Using recent generalization results \cite{arora2017generalization}, we can show that minimizing this empirical neural network distance would minimize the distance between the true distributions up to an additive error which vanishes with $n$, as follows.
\begin{theorem}  \label{thm:nn_sample}
	Under the same assumptions as in Theorem \ref{thm:gen_ub_lb}, for any class $\cF_{p,L}$ of bounded functions, which is parameterized by $u \in \reals^{p}$ and is $L$-Lipschitz in $u$, satisfying Assumption \ref{assume:invariance}, there exists a universal constant $c > 0$ such that 
	\begin{align}
	\nndff{\cF_{p,L}}{\htP}{\htQ} - \epsilon  \;\; &\leq \;\; \nndff{\cF_{p,L}}{P}{Q}  \nonumber \\
	\;\; &\leq \;\; \kappa'_\alpha \, \big( \nndff{\cF_{p,L}}{\htP}{\htQ} + \epsilon \big)\,,\nonumber
	\end{align}
	with probability at least $1 - e^{-p}$ 
	for any $\varepsilon>0$ and $n$ large enough, $
	n \;\; \geq \;\; ({c  \,p\,}/{\epsilon^{2}})\,\log\left({ p L }/{\epsilon } \right)$, where $\kappa'_\alpha =  \frac{1+\|\sum_u \alpha_{u}\|_\infty}{1-\|\sum_u \alpha_{u}\|_\infty} = \max_{i \in [n]} \frac{1+\sum_u \alpha_{ui}}{1-\sum_u \alpha_{ui}}$.
\end{theorem} 
A proof of this result directly follows from \citet[Theorem 3]{thekumparampil2018robustness} and Theorem \ref{thm:gen_ub_lb}. For more details and discussion of thes results see \citet{thekumparampil2018robustness}. Next we study some special cases of uncertainties.

\subsection{Learning from few labels}\label{sec:miss}
Assume that the true label $y$ of a sample $(x, y)$ is erased by an erasure channel with probability $\alpha^{(y)}$. As mentioned in Section \ref{sec:intro}, these missing labels could be captured by an uncertainty model with a single uncertain label $m+1$, defined by the vector $
\alpha_{m+1} = [\alpha^{(1)} \cdots \alpha^{(m)} \; 0]^T 
$
, and confusion matrix given by 
\begin{align}\label{eq:miss_confusion_matrix}
C = \diag(\ones - \alpha_{m+1})+ {\alpha}_{m+1} e_{m+1}^T\,.
\end{align}
From Theorems \ref{thm:prob_dist_ub_lb}  and \ref{thm:gen_ub_lb} we can get the following corollary.
\begin{corollary} \label{thm:miss_prob_dist_ub_lb}
	Under the same assumptions as in Theorems \ref{thm:prob_dist_ub_lb} and \ref{thm:gen_ub_lb} with $C$ given by eq. \eqref{eq:miss_confusion_matrix}, if $\bar{\alpha} = \max_y \alpha^{(y)} \neq 1$, we get,
	\begin{align}
	\tv{P}{Q}\;\;  &\leq \;\; \, 1/(1-\bar{\alpha}) \;\; \tv{\tP}{\tQ} \,, \label{eq:miss_tv_ub_lb} \\
	\;\;\js{P}{Q}\;\; &\leq\;\;  1/(1-\bar{\alpha}) \;\;   \sqrt{8\, \js{\tP}{\tQ}} \,, \label{eq:miss_js_ub_lb} \\
	\nndff{\cF}{P}{Q} \;\; &\leq \;\;  (1+\bar{\alpha})/(1-\bar{\alpha}) \;\; \nndff{\cF}{\tP}{\tQ} \,.
	\end{align}
\end{corollary}
If for all classes $y$, $\alpha^{(y)} = 1$, then RHS becomes $\infty$, which is expected since for this case labels are independent of the samples and recovery of true distribution is infeasible. As a special case, when the $\alpha$ fraction of the labels are missing uniformly at random, we have $\nndff{\cF}{P}{Q} \leq (1+\alpha)/(1-\alpha) \, \nndff{\cF}{\tP}{\tQ}$.

\subsection{Complementary labels}
Here, we assume that $\alpha$ fraction of the real class labels are changed to one of their corresponding $m-1$ complementary labels at random, i.~e.~for a real sample $(x,y)$, with probability $\alpha$ its label is changed to an uncertain label saying `$x$ is not from the class $y^{c}$' where $y^{c}$ is selected uniformly at random from $[m] \setminus \{y\}$. As discussed in Section \ref{sec:intro}, we can capture this corruption by an uncertainty model with a set of $m$ uncertain classes, $\{u_y = m+y\}_{y=1}^m$, such that $\alpha_{u_y} = \alpha [(\ones_{m} - e_y)^T \;\; 0]^T/(m-1)$, and a confusion matrix,
\begin{align}\label{eq:comple_confusion_matrix}
C = \diag(\ones - \sum_{y \in [m]}\alpha_{u_{y}})+ \sum_{y \in [m]}\alpha_{u_{y}}  e_{m+y}^T\,.
\end{align}
Again using Theorems \ref{thm:prob_dist_ub_lb}  and \ref{thm:gen_ub_lb}, we get the following guarantee.
\begin{corollary} \label{thm:comple_prob_dist_ub_lb}
	Under the same assumptions as in Theorems \ref{thm:prob_dist_ub_lb} and \ref{thm:gen_ub_lb} with $C$ given by eq. \eqref{eq:comple_confusion_matrix}, if $\bar{\alpha} = \max_y \alpha^{(y)} \neq 1$, and $\kappa_\alpha = \frac{m-1}{\alpha + (1-\alpha)(m-1)}$ and $\kappa_\alpha' = \frac{1+\alpha}{1-\alpha}$, we get,
	\begin{align}
	\tv{P}{Q}\;\;  &= \;\; \, \kappa_\alpha \;\; \tv{\tP}{\tQ} \,, \label{eq:comple_tv_ub_lb} \\
	\;\;\js{P}{Q}\;\; &\leq\;\;  \kappa_\alpha \;\;   \sqrt{8\, \js{\tP}{\tQ}} \,, \label{eq:comple_js_ub_lb} \\
	\nndff{\cF}{P}{Q} \;\; &\leq \;\;  \kappa_\alpha' \;\; \nndff{\cF}{\tP}{\tQ} \,.
	\end{align}
\end{corollary}
The multiplicative factor $\kappa'_\alpha$ can be tightened further with additional simple assumptions on the discriminator architecture.

\section{Experiments} \label{sec:expt}
\begin{table*}
	\begin{subtable}{0.5\linewidth}\centering
		{ \begin{tabular}{@{}c@{}}
				\begin{tabular}{ c  c  c  c  }
					{\bf  \#labels ($n$)} & {\bf RCGAN} & {\bf S3-GAN} \\
					\midrule
					80  &	0.977 $\pm$ 0.001	&	0.851 $\pm$ 0.014 \\ 
					60	&	0.974 $\pm$ 0.001	&	0.795 $\pm$ 0.018	\\
					40	&	0.978 $\pm$ 0.000	&	0.758 $\pm$ 0.031 \\
					30	&	0.971 $\pm$ 0.004	&	0.726 $\pm$ 0.025 \\
					20	&	0.918 $\pm$ 0.029	&	0.596 $\pm$ 0.031 \\
					10	&	0.838 $\pm$ 0.044	&	0.414 $\pm$ 0.027 \\
				\end{tabular} \\
				\midrule
				{\bf ClusterGAN} (permutation corrected): 0.901 $\pm$ 0.014 \\
		\end{tabular}}
		\caption{Generated label accuracy} \label{tab:mnist_gen_label_acc}
	\end{subtable}%
	\begin{subtable}{0.5\linewidth}\centering
		{ 	\begin{tabular}{@{}c@{}}
				\begin{tabular}{ c  c  c  c  }
					{\bf  \#labels ($n$)} & {\bf RCGAN} & {\bf S3-GAN} \\
					\midrule
					80  &	0.916 $\pm$ 0.005	&	0.880 $\pm$ 0.006 \\ 
					60	&	0.908 $\pm$ 0.005	&	0.842 $\pm$ 0.013 \\
					40	&	0.913 $\pm$ 0.007	&	0.799 $\pm$ 0.023 \\
					30	&	0.910 $\pm$ 0.009	&	0.769 $\pm$ 0.019 \\
					20	&	0.874 $\pm$ 0.024	&	0.644 $\pm$ 0.040 \\
					10	&	0.791 $\pm$ 0.042	&	 0.474 $\pm$ 0.023 \\
				\end{tabular} \\
				\midrule
				{\bf ClusterGAN} (permutation corrected): 0.855 $\pm$ 0.015 \\
		\end{tabular}}
		\caption{Label recovery accuracy} \label{tab:mnist_label_recovery_acc}
	\end{subtable}
	\caption{Average metrics ($\pm$ standard error) for RCGAN \& S3-GAN trained with MNIST dataset with very few number of labels ($n$).} \label{tab:mnist_metrics}
\end{table*}

For evaluating the empirical performance of RCGAN we consider the case of uniformly missing true class labels (Section \ref{sec:miss}) in MNIST dataset of $10$ handwritten digits \cite{mnist}. For training we use all the $70$k samples of MNIST, however only a fraction of these are labeled. We use two different metrics to evaluate the trained conditional generators: (a) generated label accuracy; and (b) label recovery accuracy. For more details on the architectures, training hyper-parameters and evaluation metrics, and more results please refer Appendix \ref{app:expt}. 

As a proof of concept, first, we show that RCGAN learns the true conditional distribution when only a significantly small fraction ($\alpha$) of the samples have labels. We see that RCGAN gets 99\%  accuracy on both metrics even when only 20\% of the samples are labeled (Table \ref{tab:mnist_gen_label_acc_fraction}). However, when $\alpha$ is below 5\% we get poor performance, which we address in the next section.

\begin{table}[h]
	\begin{center}
		\begin{tabular}{ c  c  c  c  }
			{\bf \begin{tabular}{@{}c@{}} Fraction \\ labeled ($\alpha$)\end{tabular} } & {\bf \begin{tabular}{@{}c@{}} Generated  \\ label accuracy \end{tabular} } & {\bf \begin{tabular}{@{}c@{}} Label recovery \\ accuracy \end{tabular} } \\
			\midrule
			1.0	&	0.992	&	0.924 \\  
			0.8	&	0.993	&	0.926 \\
			0.6	&	0.991	&	0.908 \\
			0.4	&	0.994	&	0.916 \\
			0.2	&	0.988	&	0.926 \\
			0.1	&	0.983	&	0.910 \\
			0.05	&	0.162	&	0.420 \\
			0.025	&	0.122	&	0.234 \\
		\end{tabular}
		\caption{{\em Generated label accuracy} and {\em Label recovery accuracy} of RCGAN trained on MNIST dataset with only an $\alpha$ fraction of samples being labeled (1 trial for each setting).}
		\label{tab:mnist_gen_label_acc_fraction}
	\end{center}
\end{table}

\subsection{Learning from {\em extremely} few labels} \label{sec:vfew-labels}
In this section we look at the case when only a very few number, $n \in \{ 10, 20 ,30, 40, 60, 80 \}$, of samples are labeled.  Since the fraction of labeled samples are extremely small we use the following modified loss function, RCGAN($\lambda$),  to boost the signal from the labeled samples.
\begin{align} \label{eq:loss_adv_vfew}
&L_{\lambda}(D, G) =  \\
&\expectD{x \sim P_{X}}{\phi\left(\D(x,e_{m+1})\right)} + \expectD{x \sim Q_{X}}{\phi\left(1 - \D(x,e_{m+1})\right)} + \nonumber \\
& \lambda\, \expectD{\substack{(x, y) \sim \\ P_{X, Y}}}{\phi\left(\D(x,y)\right)} +  \lambda\, \expectD{\substack{z \sim N \\ y \sim P_{Y}}}{\phi\left(1- \D(G(z;y), y)\right)}\,, \nonumber
\end{align}
where $\lambda  > 0$. It is easy to show that, in expectation, this loss is equivalent to the RCGAN loss when $(1+\lambda)^{-1}$ fraction of the labels are missing. Therefore, with sufficient number of samples, the above loss can recover the true conditional distributions. In our experiments, we use $\lambda=0.1$, and the first two expectations are computed with all the available real and generated samples, and the latter two expectations are computed with only the labeled real and generated sample. Note that, all the terms use the same discriminator network. 

 As a baseline, we consider the recently proposed S3-GAN \cite{lucic2019high}, which uses self(-semi)-supervised learning techniques and projection discriminator to achieve state-of-the-art image quality metrics from few labels in ImageNet dataset. We also provide the permutation corrected metrics achieved by the unsupervised ClusterGAN \cite{mukherjee2018clustergan} which learns conditional GAN from unlabeled data.
 We see that RCGAN consistently out performs S3-GAN on both the metrics (Tables \ref{tab:mnist_gen_label_acc} and \ref{tab:mnist_label_recovery_acc}). We also note that RCGAN is easier to implement than S3-GAN due to latter's pre-processing step, and S3-GAN is slower to converge.
 
In Figure \ref{fig:mnist_samples} (in Appendix \ref{app:expt}), we provide the samples generated by the RCGAN and S3-GAN architectures for $n \in \{10, 20, 30, 40\}$. In each setting, each row corresponds to a class learned by the corresponding conditional generator. We see that RCGAN produces more number of higher quality samples from the correct classes than S3-GAN which produces more number of lower quality samples from the wrong classes.

We hypothesize that this gain of RCGAN over the baselines would be more pronounced on more complex datasets such as CIFAR \cite{KH09} and ImageNet \cite{russakovsky2015imagenet}.\\


\section{Conclusion} \label{sec:conclude}
We proposed a robust conditional GAN (RCGAN) architecture which was theoretically shown to be robust to a general class of {\em uncertain labels}. This class of uncertain labels can capture a variety of label corruption models such as missing labels, complementary labels, and group memberships label. 
Further, we empirically verified its robustness on MNIST dataset when only a few labels are given.
RCGAN was able to achieve 97\% accuracy even with a few labeled examples per class.

\pagebreak

\bibliography{gan,_gan}
\bibliographystyle{icml2019}

\appendix


\section{Appendix}

\subsection{Additional theoretical results and proofs}\label{app:theory_additional}

\subsubsection{Proof of Theorem \ref{thm:prob_dist_ub_lb}}\label{app:proof_prob_dist_ub_lb}
\begin{proof}
	From \citet[Theorem 1]{thekumparampil2018robustness}, we get that,	$\tv{\tP}{\tQ} \leq  \tv{P}{Q}  \leq \| C^{-1}\|_\infty \, \tv{\tP}{\tQ}$. Next, using Woodbury matrix inversion identity \cite{henderson1981deriving} on $C$ \eqref{eq:confusion_matrix}, we can show that $C^{-1} = \diag(\ones - \sum_u \alpha_u)^{-1} (\identy - \sum_{u} \alpha_u e_{u}^T)$, which implies that $\| C^{-1} \|_\infty = \max_{i \in [m]} (1+\sum_{u} \alpha_{ui})/(1-\sum_{u} \alpha_{ui})$. We can further tighten the upper-bound by noting that $P(\cX, \{u\}_{u=m+1}^{m+\tm}) = Q(\cX, \{u\}_{u=m+1}^{m+\tm}) = 0$. Inequalities for Jensen-Shannon divergence also follow from the same reasoning.
\end{proof} 

\subsubsection{Invariance Assumption}\label{app:invariance}
 For deriving similar approximation bounds as in Theorem \ref{thm:prob_dist_ub_lb}, we make the following simple assumptions on the discriminator function class $\cF$ \cite{thekumparampil2018robustness}.  First, we define an operation $\vopsymb$ over  
a matrix $T \in \reals^{\ny \times \ny}$ and 
a class $\cF$ of functions of the form $\cX\times \reals^{\ny+\tm} \to \reals$ as 
\begin{align}
\label{eq:vop}
\vop{T}{\cF} \;\; \triangleq \;\;  \Big\{  g(x,y)  = \sum_{\ty\in[\ny+\tm]} T_{y\ty}\, f(x,\ty) \;|\; f  \in \cF   \Big\}\,.
\end{align}

\begin{assume} \label{assume:invariance}
	The class of discriminator functions $\cF$ can be decomposed into three parts  
	$\cF = \{ f_1 + f_2 + c \,|\, f_1 \in \cF_1, f_2 \in  \cF_2\} $ such that $c\in\reals$ is any constant and
	\vspace{-0.25cm}
	\begin{itemize}
		\item $\vop{T}{\cF_1} \; \subseteq \; \cF_1$, for all $\maxnorm{T} \triangleq  \max_i \sum_{j} \abs{T_{ij}} = 1$,
		\item there exists a class $\cF_2'$ of functions over $x$ such that,
		\begin{align} 
		\cF_2 \;\; =  \;\; \big\{\, \alpha \,g(x,y) \,\,|\,\, &g(x,y) = f(x) \text{ for any }  \nonumber \\ &f(x) \in \cF_2'  , \text{ and  }\alpha\in[0,1] \, \big\} \;. \nonumber
		\end{align} 
	\end{itemize}
	\vspace{-0.25cm}
\end{assume}

\subsection{Experimental details and additional results}\label{app:expt}

For the experiments in Section \ref{sec:expt}, with only $\alpha$ fraction of the samples labeled, we generate the corrupted dataset by independently labeling each sample with probability $\alpha$. We only report results from 1 trial for each of the settings. Assuming that the prior of the true classes are known, it is easy to estimate the confusion matrix \eqref{eq:miss_confusion_matrix}, which will be $C = (1-\alpha)\identy_{m+1}+ \alpha \ones_{m+1} e_{m+1}^T$.

For the experiments in Section \ref{sec:vfew-labels} with very small number of labeled samples, we allocate the labeled samples equally across the $10$ classes and within each class the labeled samples are selected uniformly at random ($\alpha = n/70000$). For each setting we provide mean and standard error over 5 trials, except for RCGAN when $n=10, 20$, for which we ran 10 trials.

For RCGAN, S3-GAN \cite{lucic2019high}, and ClusterGAN \cite{mukherjee2018clustergan} we use the same underlying discriminator and generator architectures as \citet{thekumparampil2018robustness}. 
For the modified loss \eqref{eq:loss_adv_vfew} we use $\lambda=0.1$ after a simple parameter search. For S3-GAN we use $\alpha\text{ (different from the $\alpha$ used in our paper)} = \beta=0.5$ \cite{lucic2019high}. S3-GAN uses self(-semi)-supervised pre-processing step to estimate the true labels, for which we used $\gamma=0.5$ \cite{lucic2019high}. For the pre-processing step, we use a standard CNN classifier architecture which can get 99+\% accuracy on fully labeled MNIST dataset. For ClusterGAN, we use $\beta_{\rm n} = \beta_{\rm c} = 1.0$ \cite{mukherjee2018clustergan}.
We train the RCGAN and ClusterGAN for 30 epochs, and S3-GAN for 100 epochs since it was slow to converge.

The two metrics were proposed by \citet{thekumparampil2018robustness}. Generated label accuracy is the accuracy of the generated labels, as per a pre-trained classifier with a high accuracy (99.2\%) as mentioned in \citet{thekumparampil2018robustness}. We use this classifier to predict the labels of the generated images, which are then compared with the generated labels to compute this accuracy. This is a measure of correctness of the class label ($y$) conditioning in the generator output. Label recovery accuracy is the accuracy with which the learned generator can be used to recover the true class labels of the unlabeled samples in the training data, using simple back-propagation on the conditional generator \cite{thekumparampil2018robustness}. This is a measure of the quality and coverage of the generated samples (given the generated label accuracy is high).

Since ClusterGAN is trained without any labels in an unsupervised fashion, for it we report the same metrics but after permutation correction. That is, we report the minimum metric values possible over all possible permutations of the classes learned by the conditional generator.

\begin{table}[h]
	\begin{center}
		\begin{tabular}{ c c c c c c}
			{\bf \#labels ($n$)} & {\bf S3-GAN} \\\hline
			100  & 0.725 $\pm$ 0.012 \\  
			80   & 0.673 $\pm$ 0.009  \\ 
			60   & 0.625 $\pm$ 0.010  \\  
			40   & 0.580 $\pm$ 0.017 \\  
			30   & 0.544 $\pm$ 0.018  \\ 
			20  & 0.439 $\pm$ 0.019 \\  
			10   & 0.305 $\pm$ 0.019 \\ 	
		\end{tabular}
		\caption{Average accuracy ($\pm$ standard error) of the self(-semi)-supervised classifier used in the pre-processing step of S3-GAN trained with MNIST dataset with very few number of labels ($n$).}
		\label{tab:mnist_classifier_acc}
	\end{center}
\end{table}

Finally we report the accuracy of the self(-semi)-supervised classifier from the pre-processing step of S3-GAN as a measure of the its ability to understand the true classes of the unlabeled training data. We see that the classifier has low accuracy when very few samples are labeled (Table \ref{tab:mnist_classifier_acc}), which could explain the low performance of S3-GAN when compared to RCGAN.

  \begin{figure}[h]
  	\centering
	\begin{subfigure}[b]{\textwidth}
	\includegraphics[width=0.22\textwidth]{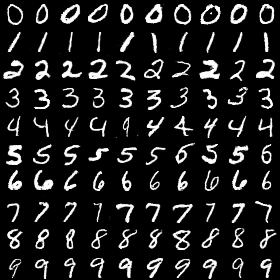}	
	\hspace{0.25em} 
	\includegraphics[width=0.22\textwidth]{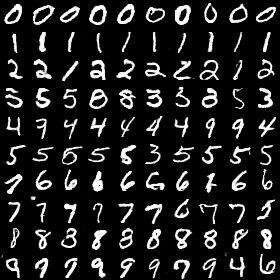}	
	\put(-190,115){RCGAN}	
	\put(-70,115){S3-GAN}	
	\put(-235,40){\rotatebox{90}{$n=40$}}
	\end{subfigure}
	\par\bigskip
	\begin{subfigure}[b]{\textwidth}
	\includegraphics[width=0.22\textwidth]{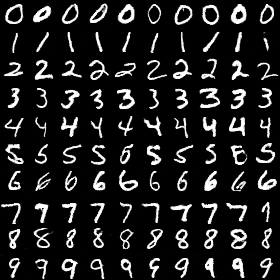}	
	\hspace{0.25em} 
	\includegraphics[width=0.22\textwidth]{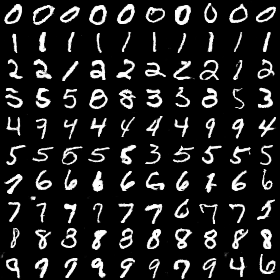}	
	\put(-235,40){\rotatebox{90}{$n=30$}}
	\end{subfigure}
	\par\bigskip
	\begin{subfigure}[b]{\textwidth}
	\includegraphics[width=0.22\textwidth]{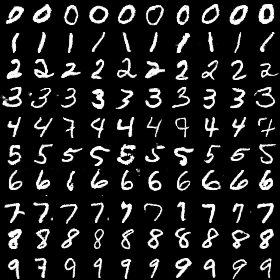}	
	\hspace{0.25em} 
	\includegraphics[width=0.22\textwidth]{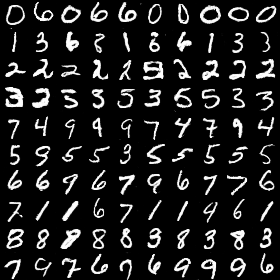}	
	\put(-235,40){\rotatebox{90}{$n=20$}}
	\end{subfigure}
	\par\bigskip
	\begin{subfigure}[b]{\textwidth}
	\includegraphics[width=0.22\textwidth]{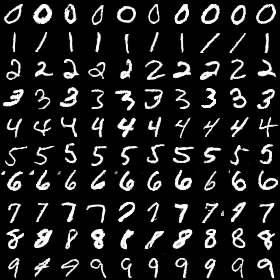}	
	\hspace{0.25em} 
	\includegraphics[width=0.22\textwidth]{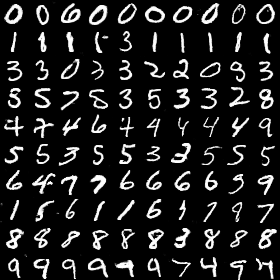}	
	\put(-235,40){\rotatebox{90}{$n=10$}}
	\end{subfigure}
	\caption{Samples generated by RCGAN and S3-GAN when trained on MNIST dataset with $n \in \{10, 20, 30, 40\}$ labels. Each row is one class as learned by the corresponding conditional generator.} \label{fig:mnist_samples}
\end{figure}

\end{document}